\documentclass[conference]{IEEEtran}
\IEEEoverridecommandlockouts
\usepackage{cite}
\usepackage{amsmath,amssymb,amsfonts}
\usepackage{algorithmic}
\usepackage{graphicx}
\usepackage{textcomp}
\usepackage{xcolor}
\usepackage{hyperref}
\def\BibTeX{{\rm B\kern-.05em{\sc i\kern-.025em b}\kern-.08em
    T\kern-.1667em\lower.7ex\hbox{E}\kern-.125emX}}
\begin{document}

\title{Raspberry Pi Bee Health Monitoring Device\\
}

\author{\IEEEauthorblockN{1\textsuperscript{st} Jakub Nevláčil, 2\textsuperscript{nd} Šimon Bilík, 3\textsuperscript{rd} Karel Horák}
\IEEEauthorblockA{
\textit{Department of Control and Instrumentation} \\
\textit{Faculty of Electrical Engineering and Communication} \\
\textit{Brno University of Technology} \\
Brno, Czech Republic \\
xnevla00@vutbr.cz, bilik@vut.cz, horak@vut.cz}
}

\maketitle

\begin{abstract}
 A declining honeybee population could pose a threat to a food resources of the whole world one of the latest trend in beekeeping is an effort to monitor a health of the honeybees using various sensors and devices. This paper participates on a development on one of these devices. The aim of this paper is to make an upgrades and improvement of an in-development bee health monitoring device and propose a remote data logging solution for a continual monitoring of a beehive.
\end{abstract}


\section{Introduction}
Honeybees are crucial for our ecosystem and also for food production around the globe. One of the main purposes of the honey bees is to pollinate flowers and plants and produce honey. The pollination of the plants plays a key role in the agriculture, because without the bees the plants could not reproduce and grow its fruits or other crops.\cite{Bee_importance} Beekeeping is a thousands of years old craft and as times goes the bees appear to be more and more endangered. Whether it is endangering by pests and diseases or today especially by pesticides from nearby agricultural fields.\cite{Pesticides} 

This is the reason why it is important for the beekeepers to be aware of the honeybees health and activity in real time and in case of an irregular bee behavior react upon this danger as soon as possible. Therefore this project aims to modify a device that collects useful data from a beehive to be able to log this data to a database where the data could be visualized using a graphical interface. Later the data will be used to determine the health and the activity of the bees using machine learning algorithms.

\section{Data acquisition device}

\subsection{Original data acquisition device}

\begin{figure}[!h]
  \begin{center}
    \includegraphics[scale=0.38]{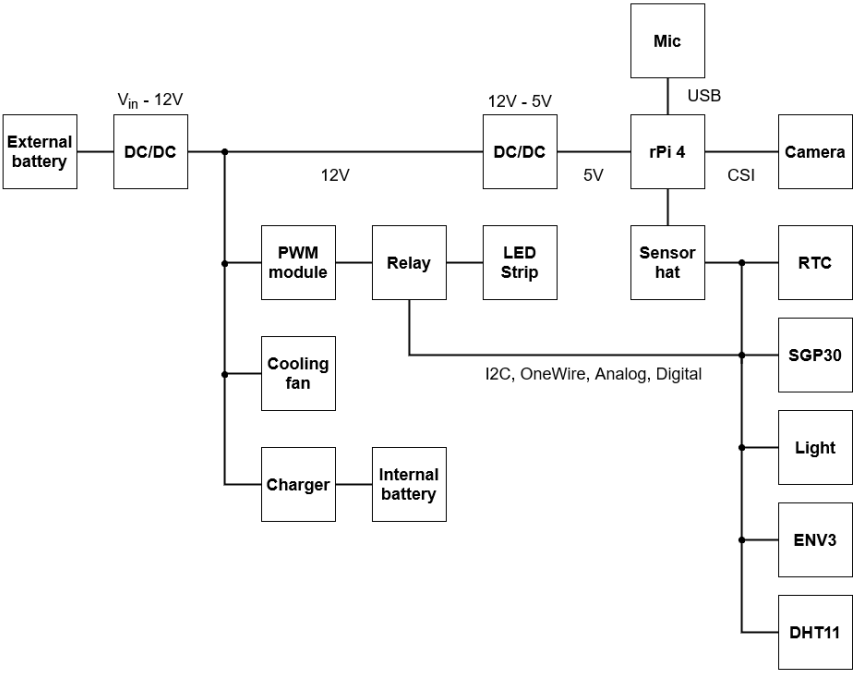}
  \end{center}
  \caption[Original data acquisition device]{Original data acquisition device}
  \label{fig:Original data acquisition device}
\end{figure}

The device for data acquisition is based on a previous work of \cite{Vcelkator_github} and \cite{Vcelkator_v1}. The original concept of the data acquisition device can be seen in the Fig.~\ref{fig:Original data acquisition device}. This device is based on a Raspberry Pi platform with a Grove Base Hat for connecting sensors to capture weather and atmospheric conditions inside and outside of the beehive. Furthermore this device also monitors the traffic of the incoming and outcoming bees using the built-in camera module and takes sound samples from the inside of the beehive with a microphone.

\subsection{Modified data acquisition device}
Although the data acquisition from the sensors and microphone usually occurs every 15 minutes (the interval can be adjusted, but there is most likely no use in collecting the data more frequently), the SGP30 sensor that measures CO2 concentration and TVOC (total volatile organic compounds) inside the beehive needs to be activated every second in order to provide correct readings. This task can be unnecessarily demanding on computing power, which can be later utilized for later implementation of the machine learning algorithms for determining health and activity of the bees.

For this purpose a Raspberry Pico was incorporated into the data acquisition chain. It was connected to the Raspberry Pi over USB port and serial communication between the two devices was established. This allows to transfer all the weather and atmospheric data from the sensors to the Raspberry Pico using the corresponding sensor Hat for this device. The Raspberry Pico is programmed in such a way that it acquires data from all the weather and atmospheric sensors every second. Then once the Raspberry Pi asks for the measured data, the Raspberry Pico calculates an average value over the inter-request period of the data from every sensor and sends them to the Raspberry Pi. The outcome of this modification is that the Raspberry Pi does not have to dedicate an entire computational thread to call the SGP30 sensor every second and can use this thread for other purposes. It should also increase reliability of the camera capturing.

\begin{figure}[!h]
  \begin{center}
    \includegraphics[scale=0.45]{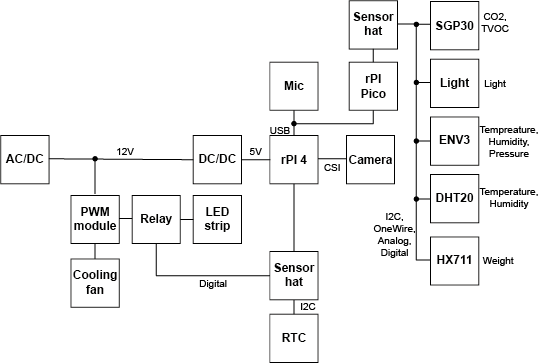}
  \end{center}
  \caption[Modified data acquisition device]{Modified data acquisition device}
  \label{fig:Modified data acquisition device}
\end{figure}

\begin{figure}[!h]
  \begin{center}
    \includegraphics[scale=0.03]{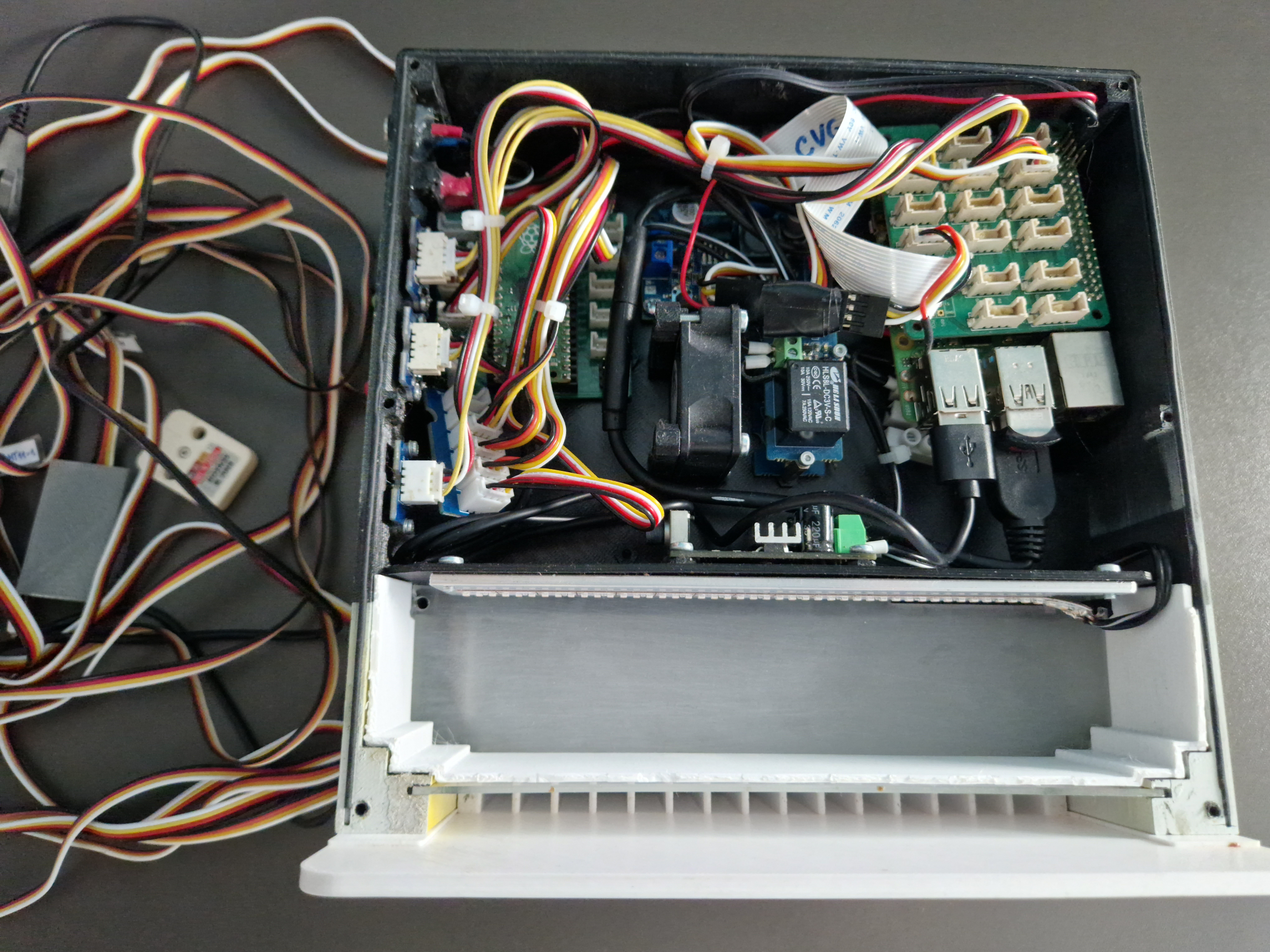}
  \end{center}
  \caption[Modified data acquisition device inside a box]{Modified data acquisition device inside a box}
  \label{fig:Modified data acquisition device inside a box}
\end{figure}

In the process of modifying, the DHT10 sensor which measures the temperature and humidity inside of the hive was updated to a newer version DHT20. This modification should bring a higher measurement accuracy and a greater measurement range. In addition to the weather and atmospheric sensors was added a HX711 amplifier with 4 load cells in full bridge configuration. This sensor will later be used for measuring the weight changes of individual beehives. A scheme of our modified data acquisition device in its final configuration can be seen from Fig.~\ref{fig:Modified data acquisition device}. In the Fig.~\ref{fig:Modified data acquisition device inside a box} we can then see a real implementation of the data acquisition device inside a box with a tunnels for the passage of the bees.

\section{Logging data to a database}

\subsection{Internet connection}

After acquiring the useful data from the hive, we want to send them over the internet to a database where we can look at the and assess the state of the hive from anywhere we like. This of course requires an internet connection. Due to the fact that beehives are usually placed on remote locations, where Wi-Fi or wired internet connection is not an option, this can be a problem. Luckily the place of deployment of our devices should have sufficient Wi-Fi coverage. In situations where a Wi-Fi connection cannot be established, solutions as GSM module or LoRaWAN could be used easily.

\subsection{MQTT messages}

The device currently stores all the acquired data from the hive on an USB flash drive. This includes sound samples from a hive, pictures of the bees and data from the weather and atmospheric sensors. For the purpose of assessing the state of health of the beehive, only the data from the weather sensors needs to be logged to the database. In future this will also include the outputs from a machine learning models that should quantify the health of the beehive and activity of the bees. Due to this fact that only a few variables needs to be transmitted to a database, the MQTT messages were chosen for this purpose.

\begin{figure}[!h]
  \begin{center}
    \includegraphics[scale=0.2]{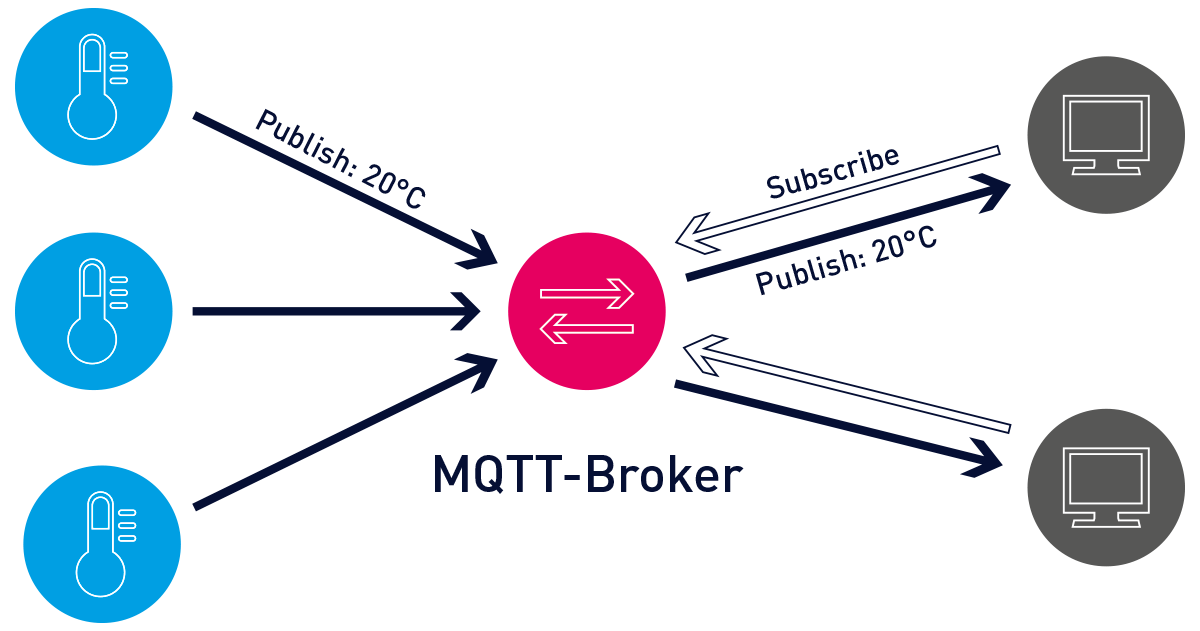}
  \end{center}
  \caption[MQTT architercture]{MQTT architercture\cite{MQTT_picture}}
  \label{fig:MQTT architercture}
\end{figure}

MQTT is a protocol used in IoT. One of the main advantages of the MQTT protocol is its lightweight design. Usually this protocol is used on a microcontrollers. This means that the MQTT protocol requires minimal computational resources and also the internet connection requirements are relatively low.\cite{MQTT}
The MQTT space consists of two kind of devices, MQTT clients and a MQTT broker. MQTT clients can either subscribe or publish messages. MQTT broker controls the flow of the messages by receiving messages from publishing clients and forwarding them to the clients that are subscribed to a particular topic. For an easier understanding an example of a MQTT network can be seen in Fig.~\ref{fig:MQTT architercture}. The MQTT broker can be hosted on a cloud or it can be installed on a local device.

In our particular case the MQTT broker will be cloud based. This solution has been assess as more suitable for our particular problem. The upside of this solution is that it doesn't require a public IP address or a VPN server on the side of the subscribed client. In conclusion our setup will consist of a Raspberry Pi data acquisition device which will be configured as publishing client, then cloud based MQTT broker and one Raspberry Pi which will be configured as a subscribing client. The subscribing Raspberry Pi will also act as a database and will run a 
web application for displaying the logged data. 

\subsection{InfluxDB database}

As mentioned above, the MQTT messages will be sent from the data acquisition device to an MQTT broker. All of the data from the sensors representing a specific period of time will be sent in a single message stored as a JSON object. The MQTT broker will the distribute every message to the subscribed clients. In our case the subscribed client will be a Raspberry Pi device where the messages will be received a processed by Telegraf, which is a data collection agent. Telegraf the takes the message sorts the data from individual sensors and saves them into an InfluxDB database. The reason the InfluxDB database is used instead of other popular databases like MySQL is because The Influx DB database is a time series database. This means that this database is designed to store a data which are associated with time. This makes it the best choice for our application of plotting the measurements from sensors over time.

\subsection{Grafana}

Once the measurements from the sensors are saved in a database, we can them plot the values from individual sensors over time. For this purpose Grafana will be used. Grafana is a web application for visualizing data and metrics. This application will also run on the Raspberry Pi next to the InfluxDB database and Telegraf and will read metrics from the sensirs directly from the database and plot them into a graphs. The Grafana application can be accessed from a web browser on a device that is connected to the same local network as the device it is installed on. Grafana can also send notifications on Slack and email. This feature could be useful once the machine learning models for recognition of bee activity and health will be implemented to warn the beekeepers about emergency situations in the hive.

\begin{figure}[!h]
  \begin{center}
    \includegraphics[scale=0.35]{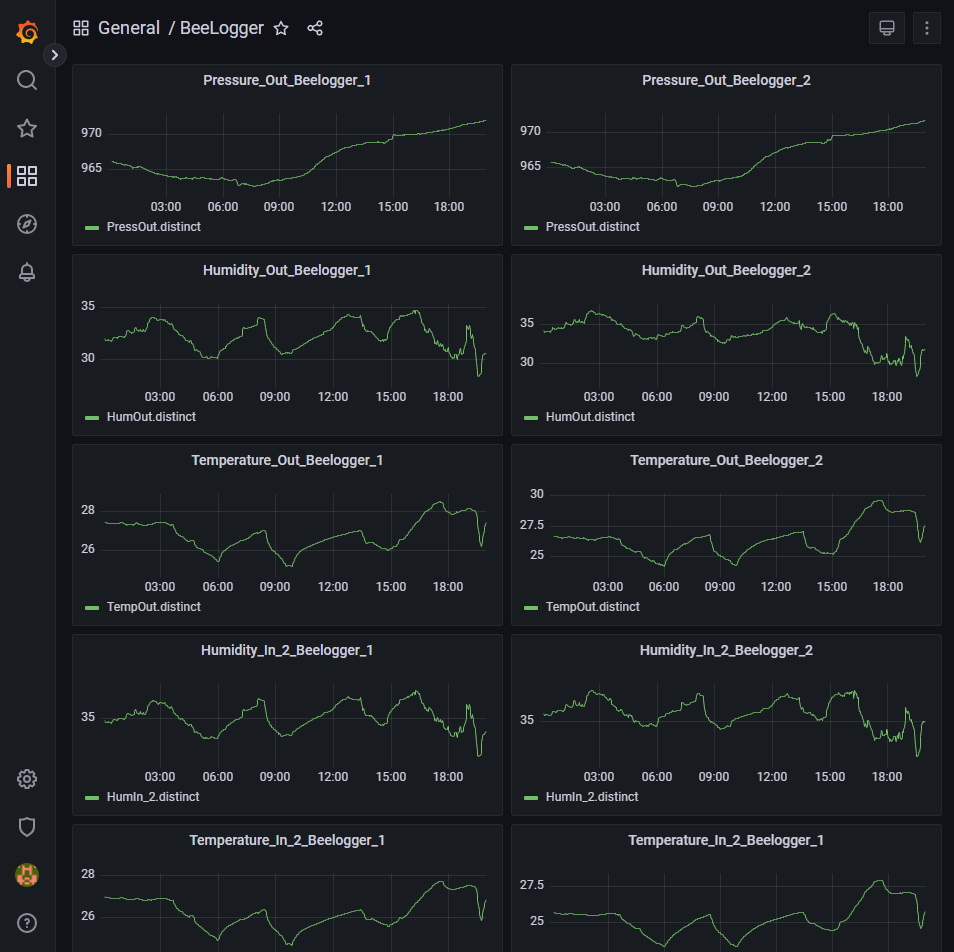}
  \end{center}
  \caption[Grafana dashboard]{Grafana dashboard}
  \label{fig:Grafana dashboard}
\end{figure}

In the Fig.~\ref{fig:Grafana dashboard} we can see a Grafana dashboard with two sets of plots. This is a data from two prototypes of the data acquisition device that was described above. This data was collected during testing phase before deployment of the devices on a real beehives.

\section{Conclusion}

In this paper we modified previous work of \cite{Vcelkator_github} to use a Raspberry Pico for collecting data from weather and atmospheric sensors to release some computational resources of a Raspberry Pi. An upgrade of a sensor was made for a newer version with better measuring accuracy and also a new weight sensor for measuring a weight changes of a beehive.

Then for an easier remote monitoring of the beehives a data logging solution was implemented. This solution starts at MQTT messages that are being sent from the data acquisition device to a cloud based MQTT broker from where the messages travel to a server with Influx DB database installed. As the last step the data is visualized using a web application Grafana, from where the visualized data can be accessed through a web browser.

In the future work an analysis of a sensor data fusion collected during a deployment on real beehives is planned. This analysis should include especially cluster analysis of a bee buzz audio samples, but also other machine learning techniques applied to both audio data and also sensor data.

\section*{Acknowledgment}
The completion of this paper was made possible by the grant No. FEKT-S-23-8451 - "Research on advanced methods and technologies in cybernetics, robotics, artificial intelligence, automation and measurement" financially supported by the Internal science fund of Brno University of Technology.

\end{document}